\documentclass[pdflatex, referee,sn-mathphys-num]{sn-jnl}

\usepackage{geometry}
\usepackage{graphicx}%
\usepackage{multirow}%
\usepackage{amsmath,amssymb,amsfonts}%
\usepackage{amsthm}%
\usepackage{mathrsfs}%
\usepackage[title]{appendix}%
\usepackage{xcolor}%
\usepackage{textcomp}%
\usepackage{manyfoot}%
\usepackage{booktabs}%
\usepackage{algorithm}%
\usepackage{algorithmicx}%
\usepackage{algpseudocode}%
\usepackage{listings}%
\usepackage{lineno}
\usepackage{dblfloatfix}

\theoremstyle{thmstyleone}%

\theoremstyle{thmstyletwo}%

\theoremstyle{thmstylethree}%

\begin{document}

\title[Article Title]{An autonomous agent for auditing and improving the reliability of clinical AI models} 

\author*[1,2,3]{\fnm{Lukas} \sur{Kuhn}}\email{lukas.kuhn@dkfz-heidelberg.de}

\author*[1,2,3,4]{\fnm{Florian} \sur{Buettner}}\email{florian.buettner@dkfz-heidelberg.de}

\affil[1]{\orgname{Goethe University Frankfurt}, \orgaddress{\city{Frankfurt}, \country{Germany}}}

\affil[2]{\orgname{German Cancer Consortium (DKTK)}, \orgaddress{\city{Frankfurt}, \country{Germany}}}

\affil[3]{\orgname{German Cancer Research Center (DKFZ)}, \orgaddress{\city{Heidelberg}, \country{Germany}}}
\abstract{The deployment of AI models in clinical practice faces a critical challenge: models achieving expert-level performance on benchmarks can fail catastrophically when confronted with real-world variations in medical imaging. Minor shifts in scanner hardware, lighting or demographics can erode accuracy, but currently reliability auditing to identify such catastrophic failure cases before deployment is a bespoke and time-consuming process. Practitioners lack accessible and interpretable tools to expose and repair hidden failure modes. Here we introduce ModelAuditor, a self-reflective agent that converses with users, selects task-specific metrics, and simulates context-dependent, clinically relevant distribution shifts. ModelAuditor then generates interpretable reports explaining how much performance likely degrades during deployment, discussing specific likely failure modes and identifying root causes and mitigation strategies. Our comprehensive evaluation across three real-world clinical scenarios — inter-institutional variation in histopathology, demographic shifts in dermatology, and equipment heterogeneity in chest radiography — demonstrates that ModelAuditor is able correctly identify context-specific failure modes of state-of-the-art models such as the established SIIM-ISIC melanoma classifier.  Its targeted recommendations recover 15-25\% of performance lost under real-world distribution shift, substantially outperforming both baseline models and state-of-the-art augmentation methods. These improvements are achieved through a multi-agent architecture and execute on consumer hardware in under 10 minutes, costing less than US\$0.50 per audit.}

\affil[4]{\orgname{Frankfurt Cancer Institute}, \orgaddress{\city{Frankfurt}, \country{Germany}}}


\maketitle


\section{Introduction}\label{sec1}

\begin{figure}[ht]
    \centering
    \includegraphics[width=\linewidth]{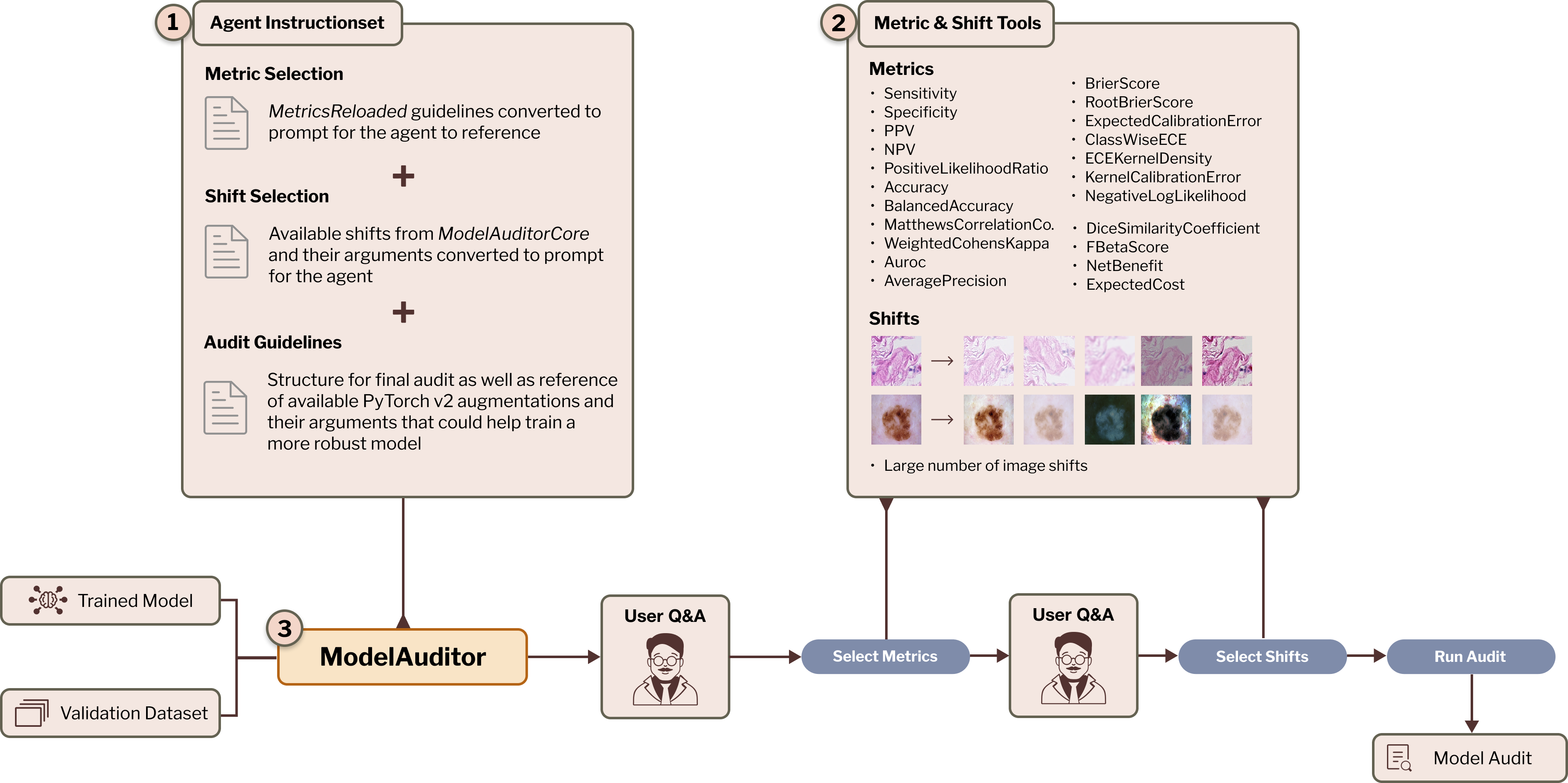}
    \caption{%
     Schematic of the \textsc{ModelAuditor} agent. 
    \textbf{(1) Agent Instructionset:} guideline texts from the \emph{MetricsReloaded} framework, the complete catalogue of synthetic distribution shifts and a template for the final report are each converted into instructions for the agent
    \textbf{(2) Metrics \& shifts tools:} extensible collection of tools containing a multitude of performance and calibration measures and distribution-shift implementations accessible to the agent 
    \textbf{(3) ModelAuditor Agent:} supplied with a trained clinical AI model and an evaluation dataset, the LLM agent autonomously chooses context-dependent, clinically relevant metrics and relevant distribution shifts and executes the audit, ultimately returning a natural-language report and actionable recommendations.
    }
    \label{fig:overview}
\end{figure}

The deployment of artificial intelligence models in clinical practice is hindered by a key challenge: models that achieve expert-level performance on benchmark datasets often fail when confronted with the subtle variations inherent in real-world medical imaging \cite{Kelly2019,Zech2018,Koh2021}. This reliability gap -- between laboratory excellence and real-world clinical safety -- has emerged as a critical barrier preventing AI from fulfilling its promise to improve healthcare delivery. Consequently, it is also reflected in evolving regulatory frameworks, including the European Union's AI Act and FDA guidance in the United States, which now mandate systematic auditing of high-risk AI systems in healthcare to ensure reliability and trustworthiness throughout the entire life-cycle of an AI system \cite{EU_AI_Act_2021,FDA_AIML_SaMD_ActionPlan_2021}. 

Consider a deep learning model trained to detect melanoma in dermoscopic images. In controlled evaluations, it may rival dermatologists in accuracy \cite{Esteva2017}. Yet when deployed in a new clinic, the same model might miss one-third of malignant lesions simply because the lighting differs from its training environment \cite{Daneshjou2022Dermatology}. Similar failures occur for other imaging modalities: chest X-ray algorithms trained at one hospital misclassify pneumonia at another due to differences in patient positioning protocols \cite{Zech2018Pneumonia}; MRI classifiers experience accuracy drops exceeding 20\% when encountering images from different scanner manufacturers \cite{Kushol2023MRI}; and histopathology models for tumor detection show marked performance degradation across institutions due to variations in tissue staining procedures \cite{Ehteshami2017,Bandi2019}.

Such distribution shifts occur frequently in clinical practice and their exact nature is usually unknown prior to model deployment. Healthcare data varies across institutions, equipment manufacturers, patient populations, imaging protocols, and time. Ensuring that a model is able to generalize across these variations via a comprehensive auditing strategy is therefore both a clinical safety imperative and regulatory requirement. 

Current approaches to auditing model reliability remain fundamentally limited. Many published models report only aggregate metrics such as area under the receiver operating characteristic curve (AUROC) or accuracy for in-distribution scenarios \cite{Maier_Hein_2024,FDA2019}, which can mask substantial performance drops in real-world deployment scenarios or patient subpopulations. Even well-intentioned practitioners lack standardized tools for systematically probing their models' failure modes before deployment \cite{disalvo2024medmnistccomprehensivebenchmarkimproved, navarro2021evaluating}.\\
A related line of work attempts to harden models during training by synthetically enriching the training data \cite{cubuk2019autoaugmentlearningaugmentationpolicies}, but suffers from three critical shortcomings. First, they require practitioners to anticipate which distribution shifts will occur -- a challenging task requiring expert domain knowledge as well as expert AI expertise given the domain-specific nature of real-world deployment environments. Second, augmentation policies that improve robustness in one imaging modality may harm performance in another \cite{goceri2023medical}. Third, they provide no mechanism for understanding \emph{when} a trained model fails under distribution shift, \emph{why} it fails and \emph{how} to mitigate these failures. Taken together, an auditing framework that quantifies the reliability of a trained AI model by comprehensively anticipating potential distribution shifts that may occur throughout its lifecycle and also explains failures in clinically interpretable and actionable terms is needed for closing the reliability-gap.

\noindent
The recent emergence of large language models (LLMs) as autonomous agents offers a transformative opportunity to address these challenges \cite{liu2024medchainagent,lu2024clinicalrag}. When augmented with specialized tools, LLM agents have demonstrated remarkable capabilities in medical contexts: generating radiology reports with iterative self-refinement \cite{zeng2024enhancingllmsimpressiongeneration}, performing sequential clinical decision support \cite{liu2024medchainagent}, and providing treatment recommendations grounded in medical literature \cite{lu2024clinicalrag}. These examples illustrate that LLMs, when orchestrated as agents, have the potential to also effectively audit the reliability of medical AI systems.


Here, we present ModelAuditor, an autonomous agent that conducts a comprehensive reliability evaluation of trained AI models that enables practitioners to understand failure modes and implement corrective measures. Given any trained clinical AI vision model, ModelAuditor conducts a comprehensive audit by: (i) engaging practitioners in natural conversation to understand the clinical context and deployment environment; (ii) automatically selecting statistically sound evaluation metrics that capture clinically meaningful failures; (iii) simulating clinically relevant distribution shifts tailored to the specific clinical context and deployment environment; and (iv) generating interpretable reports that explain \emph{how much} performance degrades, as well as \emph{why} it fails and \emph{how} to fix it.

Unlike existing approaches that treat robustness as a monolithic property, ModelAuditor recognizes that clinical relevance depends on context. A dermatology model intended for teledermatology must be robust to smartphone camera variations, while a histopathology model for multi-institutional trials must handle diverse staining protocols. By tailoring its analysis to each specific use case, ModelAuditor provides actionable insights that directly translate to better model reliability and regulatory compliance.

We demonstrate ModelAuditor's capabilities across three complementary clinical scenarios that exemplify common distribution shift challenges: inter-institutional variation in histopathology, demographic shifts in dermatology, and equipment heterogeneity in chest radiography. In each case, ModelAuditor (i) identifies critical failure modes missed by conventional evaluation and (ii) provides targeted re-training strategies that recover 15–25\% of lost performance during deployment -- substantially outperforming both baseline models and state-of-the-art strategies for data augmentation. Crucially, these improvements are achieved through a conversational interface that executes on consumer hardware in under 10 minutes, and costs less than US\$0.50 per audit.


\section{Results}\label{sec2}

ModelAuditor operates through a multi-agent architecture (Fig.~\ref{fig:overview}; section \ref{sec:tooling}) where specialized sub-agents debate to select appropriate metrics and clinically relevant distribution shifts, guided by the MetricsReloaded framework \cite{Maier_Hein_2024} and specific clinical deployment scenarios.

\begin{figure}[ht]
    \centering
    \includegraphics[width=\linewidth]{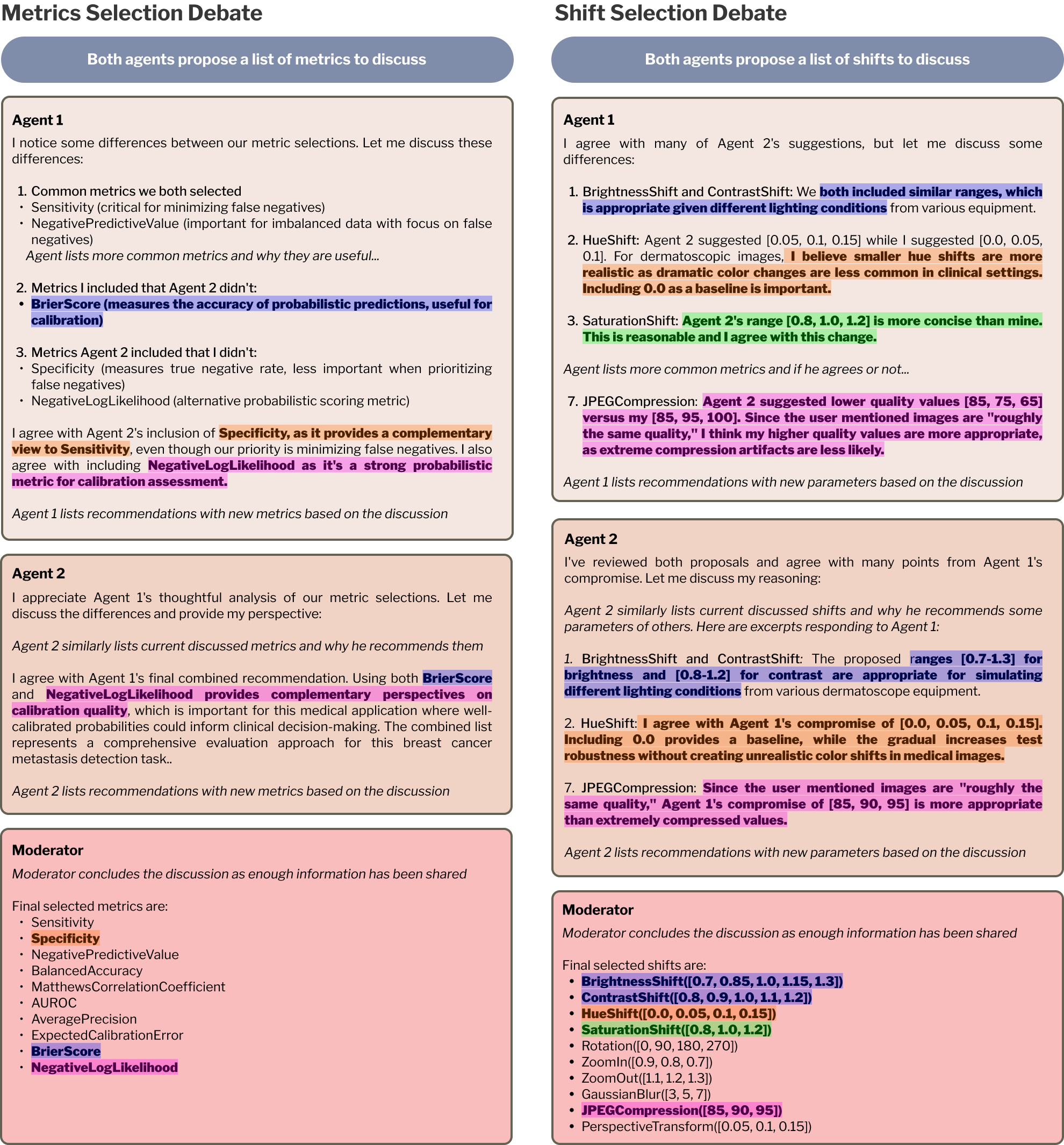}
    \caption{Two specialized sub-agents engage in a rapid debate to nominate evaluation metrics that best capture clinical risk and select distribution shifts that mimic model- and task-specific real-world variability. They engage in a discussion about the usefulness of each others' selected metrics, shifts, and their parameters to determine the final selection together \cite{du2023improvingfactualityreasoninglanguage}. Some of the debate is color coded to show consistency between the debate and selected metrics.}
    \label{fig:multiagentdebate}
\end{figure}

To demonstrate ModelAuditor's capabilities, we first tested it on a real-world failure case: the SIIM-ISIC melanoma classifier \cite{degrave_auditing_2025}, which achieved state-of-the-art performance in the ISIC 2020 challenge. 
When given the clinical context for auditing the SIIM-ISIC melanoma classifier — "We're deploying this model in a teledermatology setting with smartphone cameras across rural clinics" — the agent automatically reasoned through the implications. Through structured debate between proposer and critic sub-agents (Fig.~\ref{fig:multiagentdebate}), it identified that smartphone imaging would introduce specific challenges: variable zoom levels, inconsistent lighting, compression artifacts, and diverse backgrounds. The agent then selected evaluation metrics that captured these specific risks (sensitivity for catching melanomas, calibration for trustworthy predictions) rather than defaulting to generic accuracy measures. A comprehensive quantification of these metrics under the model- and use-case specific challenges uncovered three clinically relevant weaknesses (Fig.~\ref{fig:finalaudit}).

First, ModelAuditor identified that the model's sensitivity to melanoma detection was sensitive to routine lighting variations — dropping from 0.59 to 0.47 with modest brightness changes. In practical terms, this means that a melanoma photographed under typical examination room lighting rather than the idealized conditions of the training set has a one-in-three chance of being missed. Second, the model  performs best with a slight zoom with centered lesions, which is seldom the case in clinical practice. Third, the model systematically made predictions with overconfidence, with an expected calibration error of 0.11; that is, its confidence scores did not reflect the actual risk well, resulting in wrong predictions made with high confidence — which could lead clinicians to trust erroneous predictions.

To translate these findings into accessible and actionable insights, ModelAuditor's conversational interface translated each metric into concrete patient safety implications: "Under typical clinic lighting, this model would miss every third melanoma. The confidence scores it provides are dangerously misleading and should not guide clinical decisions."

To validate the agent's hypotheses we explored the training dataset and were able to confirm that lesions are typically well lit, in-center and without any background noise that distract from the classification task. We next tested the SIIM-ISIC melanoma classifier on unseen test data from different hospitals - the Fitzpatrick17 dataset, which consists of images that are not well cropped, with lesions being off-center, varying lighting and contrast between samples as well as frequent background noise like hair, ears and clothing. As anticipated by ModelAuditor, the SIIM-ISIC melanoma classifier fails to correctly predict malignant samples -- the AUROC of the model drops over 15\% when tested on this out-of-distribution dataset (\cite{degrave_auditing_2025}). The root causes of this real-world generalization failure were anticipated by ModelAuditor, illustrating the actionable and accurate nature of the audits.

\subsection{ModelAuditor autonomously orchestrates complex multi-tool workflows} \label{sec:tooling}

Conventional robustness studies require hand-coded test suites, ad-hoc metric choices, and expert interpretation of failure modes, steps that sharply limit their adoption in day-to-day clinical AI development \cite{disalvo2024medmnistccomprehensivebenchmarkimproved, degrave_auditing_2025}.  ModelAuditor replaces this fragmented workflow with a single, conversational interface powered by a large-language-model (LLM) agent with a suite of tools at it's disposal. The auditor is augmented with a hand-tuned instruction set based on the MetricsReloaded guidelines by Maier-Hein et al. \cite{Maier_Hein_2024}, a list of the available tools and how to use them and a guide on how to structure the final audit. 

\begin{figure}[ht]
    \centering
    \includegraphics[width=\linewidth]{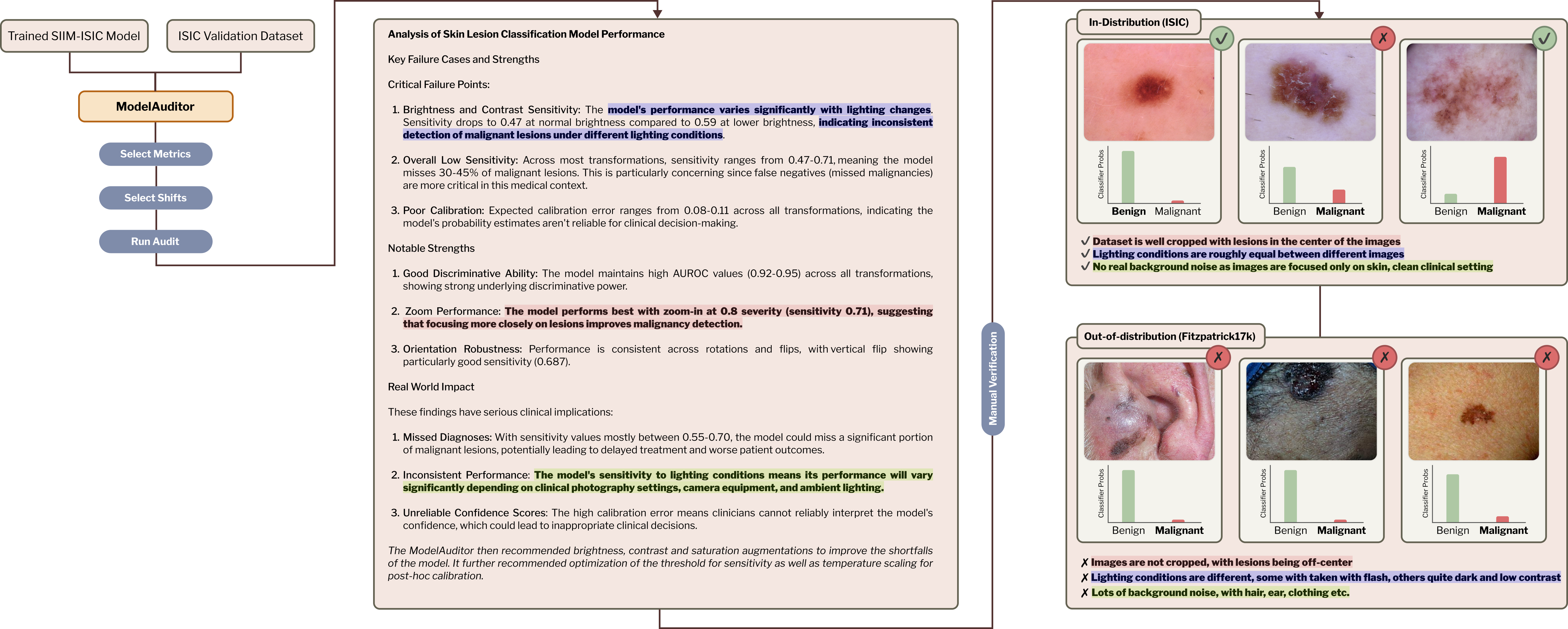}
    \caption{ModelAuditor finds multiple clinically relevant weaknesses in its SIIM-ISIC audit: \textbf{(Blue)} The model performance drops under certain lighting changes. \textbf{(Red)} SIIM-ISIC works best with the lesion in focus but struggles with zoomed-out images \textbf{(Green)} The model performance will vary significantly if used in clincal practices with different photography settings and equipment. We compare these findings with the training data (ISIC19) and an unseen out-of-distribution dataset (Fitzpatrick17) with a significantly lower AUROC and are able to confirm all hypotheses (rightmost panels). The correct label is highlighted in bold.}
    \label{fig:finalaudit}
\end{figure}

The practitioner begins by calling the command line interface with a trained vision model together with a representative in-distribution dataset (e.g. a small subset of the training dataset).  Through a short series of clarifying questions, the auditor agent elicits the clinical task (e.g.\ binary metastasis detection versus multi-label chest-pathology screening), the relevant operating point, and deployment constraints such as differences in scanner vendor, color calibration, or patient-demographic mix.  Using the answers, following Du et al. \cite{du2023improvingfactualityreasoninglanguage} two specialized sub-agents engage in a rapid debate to (i) nominate evaluation metrics that best capture clinical risk and (ii) select distribution shifts that mimic real-world variability. Once metrics and shifts are fixed, the agent calls an internal tool that executes hundreds of perturbation–evaluation cycles on a subset of the data.  Results are streamed back in natural language and distilled into an executive report that highlights  failures modes as well as strengths of the model (see Figure \ref{fig:finalaudit}). Afterwards, the user can interrogate the agent, \emph{“What are the clinical implications of the saturation shift failures?”}, \emph{“Which distribution shifts matter most?”}, and receive mitigation advice grounded in the audit evidence.  In this way, ModelAuditor implements reliability analysis as an interactive feedback loop that practitioners can employ as final audit before model deployment but also during routine model development.


\subsection{Targeted augmentations recover lost performance under distribution shift}
\label{sec:ood_results}

\begin{figure}[ht]
    \centering
    \includegraphics[width=\linewidth]{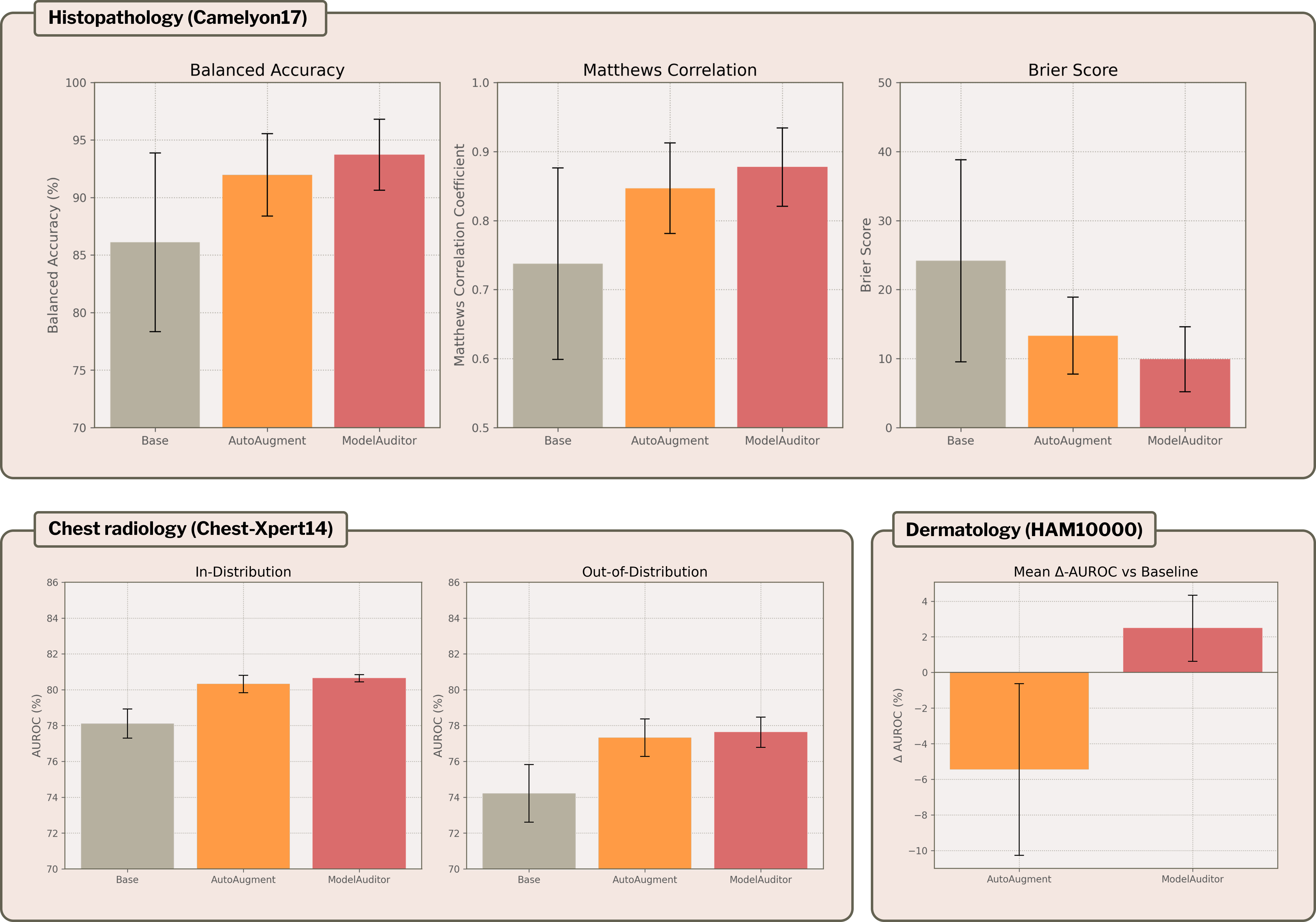}
    \caption{ \textbf{(1) Histopathology results} measured on data from hold-out hospitals (Camelyon17) that is out-of-distribution due to sampling bias. ModelAuditor significantly outperforms the Baseline and AutoAugment in all metrics. \textbf{(2) Chest radiography results}  ModelAuditor outperforms the baselines both in-distribution and out-of-distribution in terms of AUROC. \textbf{(3) Dermatology results} AutoAugment harms baseline performance on test data with demographic shift from Australia while ModelAuditor improves it due to it's targeted augmentations.}
    \label{fig:results}
\end{figure}

To systematically assess how the ModelAuditor's mitigation recommendations - grounded in its audit - improve model reliability, we applied it on three complementary settings that mirror common sources of clinical distribution shift: inter-institutional variation in histopathology, equipment differences in radiology, and demographic shifts in dermatology.


\paragraph{Inter-institutional histopathology}
To investigate the challenge of generalizing across hospitals with different staining protocols, we trained four widely adopted backbone architectures (ResNet-50/152 \cite{he2015deepresiduallearningimage}, VGG-19 \cite{he2015deepresiduallearningimage} and ViT \cite{Dosovitskiy2021}) for metastasis detection based on histopathology slides from 4 hospitals, setting aside a fifth hospital with distinct protocol (Camelyon17 dataset with WILDS split, \cite{Koh2021WILDS}).
 ModelAuditor selected (among others) balanced accuracy and Matthews correlation coefficient (MCC) as primary metrics, together with Brier score to quantify calibration - accurately abstracting a task-specific problem fingerprint and mapping it to the correct set of metrics specified within the Metrics Reloaded framework.  Across architectures, deployment of a base network led to a significant loss in accuracy (AUROC dropping 7\% on average) and calibration (Brier score 22\% higher on average) on the held-out test hospital. Using ModelAuditor's targeted augmentations — specifically addressing stain variation and tissue preparation differences — in a retraining step, recovered up to 15\% of the lost accuracy while simultaneously improving calibration, as measured by Brier score and expected calibration error (ECE).
 In contrast, choosing generic augmentations via AutoAugment for re-training, yielded consistently smaller improvements in model reliability.

\begin{figure}[ht]
    \centering
    \includegraphics[width=\linewidth]{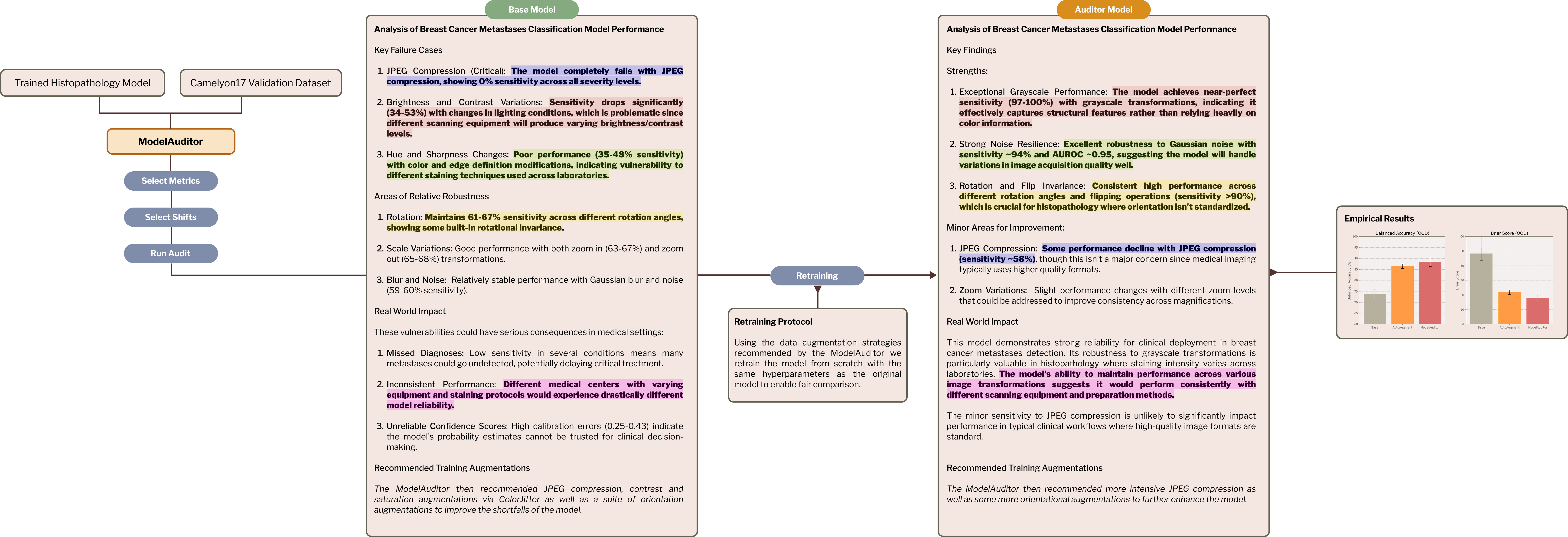}
    \caption{ \textbf{Comparison of Baseline and Audited Model:} After the original audit for the baseline ViT we reran the audit to verify that ModelAuditor can reliable identify high performance in models. Color-coded are performance comparisons between base and agent-improved model. We empirically validate the agents on an out-of-distribution held out hospital (rightmost panel).}
    \label{fig:comparison}
\end{figure}

We next assessed wether ModelAuditor could recognize when its recommendations had been successfully implemented. To this end, we used a fresh instance of ModelAuditor without any knowledge of the baseline or previous audit to audit the Vision Transformer model after retraining based on the initially suggested augmentations.
The agent identified the now high performance of the model under various variations and argued that the model now can "perform consistently with different scanning equipment and preparation methods" (see Figure \ref{fig:comparison}). We empirically verified this result and confirmed the agent's audit of the re-trained model: the model improved under all synthetic distribution shifts and on the real-world test data from a different hospital (out-of-distribution; see Table \ref{tab:camelyon17}).

\paragraph{Sampling bias in chest radiography}
Chest radiograph datasets differ markedly in disease prevalence, acquisition protocol and scanner vendor \cite{doi:10.1073/pnas.1919012117,seyyed2021underdiagnosis}.  We next tested weather ModelAuditor is able to (i) identify such sampling bias as potential issue for model deployment and (ii) is able to propose suitable mitigation strategies. To this end, we trained the four backbones for multi-label pathology detection on the CheXpert dataset \cite{Irvin2019CheXpert} containing 224,316 samples collected from 65,240 patients with 14 thoracic pathologies including Pneumonia, Edemas and Fractures. 
ModelAuditor identified that models would face different X-ray equipment and acquisition protocols when deployed in different hospitals. Using its recommendations such as randomized geometric transformations, specific color-jitter and gaussian blur as augmentations during a re-training phase, maintained robust performance in the in-distribution setting, while consistently improving AUROC on data from a different environment with real-world differences in disease prevalence and acquisition centers (ChestX-ray14 data \cite{Wang2017ChestXray8}). Notably, these gains were consistently larger compared to re-training with AutoAugment.

\paragraph{Demographic shift in dermoscopy}
To examine how well ModelAuditor can quantify and improve model realiability under demographic shifts, we partitioned HAM10000 \cite{Tschandl2018HAM10000}, a dataset for detecting pigmented skin lesions, by clinic of origin: images from the Vienna Dermatology Department formed the training set, whereas those from an Australian practice served exclusively as test data, reflecting real-world variation in imaging devices, illumination and patient skin tone. Notably, generic AutoAugment resulted in worse AUROC and Brier score on the Australian test data in comparison to even the base model, confirming that one-size-fits-all augmentation can misalign with specific demographic shifts and even hurt model generalizability \cite{goceri2023medical}.  In contrast, ModelAuditor’s custom augmentations exceeded baseline accuracy for every architecture while improving Brier score by up to 25\% (Table~\ref{tab:ham10000}).

Taken together, mitigation strategies identified by the agent based on its audit successfully protects against failures to generalise to real-world variation of these models while being interpretable for practitioners. 

In all three scenarios the selected metric sets exactly matched those a domain specialist would recommend, verified by a manual review against the MetricsReloaded checklist.  This agreement underscores that the agentic debate reliably translates free-text task descriptions into rigorous, context-appropriate evaluation criteria, laying a principled foundation for the performance results.

Across all datasets and backbones, ModelAuditor recovered most of the performance lost under distribution shift relative to in-distribution evaluation and similarly reduced calibration error compared with AutoAugment.  Crucially, these improvements arose from auditor-specific, context-aware interventions rather than stronger regularization or longer training, underscoring the value of an interactive agent that identifies precise failure modes and suggests targeted mitigation strategies.  The resulting robustness gains translate directly to clinical reliability: fewer missed metastases, more consistent chest-pathology triage across hospitals, and equitable skin-lesion detection across imaging devices.  In sum, ModelAuditor provides a principled pathway to narrow the persistent gap between benchmark excellence and real-world reliability.

\subsection{ModelAuditor delivers fast, low-cost and hardware-light auditing}
\label{sec:efficiency}

A practical robustness tool must fit the time, compute and budget constraints of clinical AI development \cite{kapoor2024aiagentsmatter}. To that end, ModelAuditor is implemented as a single, high-capacity  agentic system (based on Claude 3.7 Sonnet) that launches a short multi-agent debate only for metric and shift selection; all subsequent reasoning, explanation and Q\&A run through the same dialogue thread.  The complete audit, including clarifying questions, shift simulation, natural-language report and follow-up queries, costs less than USD \$0.50.  Model choice is configurable via a single line of code, so laboratories with institutional licenses for alternative models can swap engines without adjusting the surrounding workflow or local models can be utilized for maximum data privacy. 
\\
The end-to-end wall-clock time is dominated by user response latency.  On a standard 2024 MacBook Pro M3 (10-core CPU, 16 GB RAM) we completed all experiments in 5–10 minutes per model, including shift generation and metric computation.  Taken together, these results show that robust out-of-distribution auditing does not require specialized GPU servers, making it feasible for resource-constrained practitioners to comprehensively audit clinical AI models before deployment.

\section{Discussion}\label{sec4}

Robustness to real‐world distribution shift is recognized as a missing link between benchmark-level performance and clinical impact in medical AI \cite{su2024navigatingdistributionshiftsmedical, choi2023translating}, yet current auditing workflows remain slow, expensive and prohibitively specialized \cite{disalvo2024medmnistccomprehensivebenchmarkimproved, degrave_auditing_2025}. By distilling metric selection, shift simulation, failure analysis and mitigation into a single LLM-driven conversation, ModelAuditor implements reliability evaluation as a quality-control step that a practitioner can execute on a laptop in under ten minutes and for less than half a US dollar.  The agent’s dialogue not only quantifies how a model might fail outside its controlled training setting, but also explains why it fails and how to fix it, thereby closing the loop from model diagnosis to model reliability.

Three observations underscore the practical value of this approach.  First, the agent chose metrics that fit each task: sensitivity, specificity and calibration for histopathology (to avoid missed metastases); subset accuracy, Hamming loss and label-wise F\textsubscript{1} for chest X-rays (matching radiologists’ “all findings” rule); and discrimination plus calibration for dermoscopy (to prevent skin-tone bias). The resulting panels overlapped 100\% with the task-specific MetricsReloaded \cite{Maier_Hein_2024} checklist, showing that plain-language task descriptions can indeed be translated into rigorous evaluation metrics.

Second, the agent's targeted interventions consistently recovered predictive power and uncertainty calibration  lost under distribution-shifts, while being lightweight and targeted in its proposed augmentations. Broad ill-chosen transformations can harm performance under distribution shifts in comparison to the baseline models as been observed by Goceri et al. \cite{goceri2023medical}. In contrast, ModelAuditor with it's specific task-dependent augmentations decreases the gap between in- and out-of-distribution performance.

Third, the agent's narrative "diagnostic notes" translated abstract robustness metrics into concrete clinical risks and specific mitigation strategies. Such explanations make the aufit results interpretable and easy and actionable.

The resource profile further lowers the adoption barrier. By limiting multi-agent debate to metric and shift selection and performing all follow-up reasoning in a single Claude-3.7 Sonnet thread, a typical audit costs less than \$0.50 at current pricing.  

Our study nonetheless has limitations. The framework is currently vision-only; multimodal models that fuse imaging with text or genomics will require an expanded metric ontology and new shift generators.  While we did not observe metric mis-selection, agentic decision quality is ultimately bounded by the underlying LLM; weaker or cost-optimized models might underperform.  Finally, simulated shifts (even a large catalog) can never be exhaustive;  continuous monitoring during deployment remains an important complementary requirement for safe and efficient adoption of clinical AI models.

In summary, by implementing reliability evaluation as an interactive dialogue, ModelAuditor lowers the technical and financial hurdles to comprehensive auditing of clinical AI models.  Its task-aware metric selection, interpretable failure explanations and laptop-scale efficiency collectively facilitate actionable audits of clinical AI models.

\section{Methods}\label{sec3}

\subsection*{Agentic framework}

The entire auditing workflow is driven by a single large-language-model agent, Claude 3.7 Sonnet (200k-token context), queried at temperature 0.5.  Three lightweight instruction prompts shape the model’s behavior at different phases of the audit.  \emph{Phase 1} supplies the model with a concise rubric distilled from the \textit{MetricsReloaded} guideline grid \cite{Maier_Hein_2024}; the rubric instructs the agent to decide, in debate form \cite{du2023improvingfactualityreasoninglanguage}, which evaluation metrics best reflect the clinical task just described by the user.  \emph{Phase 2} presents a catalogue of twenty-two clinically motivated perturbations (e.g.\ stain-mixing, JPEG artefacts, small patient rotations) and asks the agent, again via proposer–critic dialogue, to select the subset most representative of the intended deployment environment.  \emph{Phase 3} provides a response template that the agent must follow when writing the final narrative and requests a structured explanation of failure modes and remediation steps. The backbone model can be swapped for any GPT- or open-weights alternative by editing a single constant in the entry-point script; no other part of the framework requires modification.

When an audit begins the agent elicits the problem fingerprint specifying binary, multi-class or multi-label classification; class imbalance; tolerance for false positives versus false negatives; and calibration requirements, through a short series of clarifying questions.  Following Du et al. \cite{du2023improvingfactualityreasoninglanguage} two ephemeral sub-agents, Proposer and Critic, then engage in a deterministic dialogue: the Proposer nominates a metric set, the Critic highlights omissions, and the core agent mediates until “no further objections" are raised. The accepted metrics are emitted as a single XML-style "metric" tag.  In all experiments the automatically chosen set overlapped 100\,\% with independent domain-expert selections and we found that accuracy and interpretability are increased by the multi-agent debate.

A second debate mirrors the same proposer–critic–mediator protocol.  Here the agent issues a deployment questionnaire covering vendor of medical imaging devices, image-compression practices, illumination variability and patient demographics.  Candidate perturbations and parameter ranges reflect values reported in the imaging‐robustness literature.  Each final choice is returned as a parameterized XML-style "shift" tag, producing a reproducible audit specification that fully documents the simulated distribution shifts.

The agent then uses tool calls to add the selected metrics and shifts to a main audit object of the core ModelAuditor package, that implements a large variety of metrics and shifts (see left side of Figure \ref{fig:overview}), and to start the audit. Perturbations are implemented with TorchVision v2 transforms and applied to a stratified (default)  10\% sample of the dataset to cap runtime.  Metric computation wraps the open-source MetricsReloaded package, ensuring one-to-one correspondence with the debate output.  Batched evaluation runs in parallel across CPU cores; a complete audit finishes in 5–10 min on a 2024 MacBook Pro M3 (10-core CPU, 16 GB RAM).  The ModelAuditorCore returns a DataFrame and a markdown summary table, both passed back to the LLM for conversational interpretation.  Typical audits consume 75k input and 10k output tokens, for a median marginal cost of USD \$0.50 at current API rates.

\subsection*{Datasets}

\paragraph{WILDS Camelyon17}
Camelyon17 comprises whole-slide histopathology images from five pathology laboratories; each slide is annotated at the pixel level for the presence or absence of lymph-node metastases.  The WILDS split \cite{Koh2021WILDS} uses slides from three institutions for training, one for validation and reserves the fifth for out-of-distribution  testing, thereby capturing the inter-laboratory stain and scanner variability known to erode real-world performance. We only used the data of two of the hospitals in the training split and used the third for our in-distribution results. We framed the task as binary patch-level classification,metastatic vs.\ benign tissue,because false negatives carry the highest clinical risk and because balanced accuracy, MCC and calibration can be directly interpreted by pathologists making treatment decisions.

\paragraph{CheXpert and Chest-Xray14}
CheXpert is a large-scale repository of 224,316 frontal and lateral chest radiographs labelled for 14 thoracic pathologies with expert adjudication of uncertainty \cite{Irvin2019CheXpert}.  We trained models on CheXpert and evaluated OOD performance on Chest-Xray14, an independent collection of 112,120 images sourced from different hospitals, scanners and acquisition protocols \cite{Wang2017ChestXray8}.  This pairing isolates the “sampling-bias" shift: disease prevalence, view position and image contrast differ markedly between the two datasets, mirroring the challenges of deploying triage models across healthcare systems.  The multi-label nature of the task motivates the auditor’s choice of subset accuracy, Hamming loss and label-wise precision/recall as primary metrics.

\paragraph{HAM10000}
HAM10000 contains 10,015 dermatoscopic images spanning seven categories of pigmented skin lesions \cite{Tschandl2018HAM10000}.  Images originate from two primary sources: the ViDIR group at the Medical University of Vienna, Austria (“vidir\_modern") and the private practice of Dr.\ Cliff Rosendahl in Queensland, Australia (“rosendahl").  To probe demographic and device-related drift we trained exclusively on the Vienna subset and treated the Australian clinic as OOD test data.  This split captures variations in dermatoscope hardware, illumination and, crucially, patient skin tone, providing a realistic proxy for equity-related deployment challenges.  Because all available Vienna images are required to train competitive classifiers, we report only OOD metrics (AUROC and Brier score) for this dataset.

\subsubsection*{Architectures \& Training}

We evaluated four widely adopted backbone families, two convolutional residual networks, one deep convolutional baseline and one transformer, using the off-the-shelf ImageNet-pretrained weights available in \texttt{torchvision}.  

\begin{itemize}
    \item \textbf{ResNet-50}~\cite{He2016}.  This 25.6-M parameter network with a 4-stage residual hierarchy serves as our baseline; its popularity in biomedical imaging makes it a useful point of reference.  
    \item \textbf{ResNet-152}~\cite{He2016}.  A 60.2-M parameter, 152-layer variant that explores whether depth confers additional robustness under distribution shift.  
    \item \textbf{VGG-19}~\cite{Simonyan2015}.  A 143-M parameter architecture comprising 16 convolutional and three fully connected layers; although superseded in accuracy by residual models, VGG remains common in clinical-AI codebases.  
    \item \textbf{ViT-B/16}~\cite{Dosovitskiy2021}.  A pure transformer that tokenises each \(224\times224\) image into \(16\times16\) patches and processes them with 12 transformer blocks; we use the ImageNet-21k–pretrained, ImageNet-1k–fine-tuned checkpoint.  
\end{itemize}

All models were fine-tuned under the same regime to isolate the effect of auditor-recommended augmentations.  We employed the Adam optimizer (\(\beta_1{=}0.9\), \(\beta_2{=}0.999\)) \cite{kingma2017adammethodstochasticoptimization}; learning rates of \(1\times10^{-3}\) for the ResNets and ViT, and \(1\times10^{-4}\) for VGG-19 to prevent unstable loss escalation.  A one-cycle schedule with two warm-up epochs followed by cosine decay over eight further epochs controlled the learning rate.  Loss functions matched dataset label structure: Cross-Entropy loss for Camelyon17 and HAM10000, and Binary-Cross-Entropy loss for the multi-label CheXpert task.  Each model was fine-tuned for ten epochs and repeated with three independent random seeds (\(seed\in\{1,2,3\}\)); averages and standard errors are reported throughout.

\subsubsection*{Benchmark augmentation policies}

All experiments apply a single, dataset-agnostic resizing operation that scales every image to \(224\times224\) pixels; this step is common to all three conditions, Base, AutoAugment and ModelAuditor, and therefore does not influence relative comparisons.

\paragraph{Base}
The base configuration performs no augmentation beyond the obligatory resize.  This choice mirrors many published clinical-AI pipelines, in which practitioners rely on large pretrained weights and avoid additional transforms to not introduce artefacts.

\paragraph{AutoAugment}
For the AutoAugment baseline we used the “ImageNet” policy from Cubuk et al. ~\cite{cubuk2019autoaugmentlearningaugmentationpolicies}.  The policy comprises 25 sub-policies, each a pair of stochastic operations (e.g.\ Posterize followed by Rotate) applied with learned probabilities and magnitudes.  We implemented the policy via the reference torchvision wrapper, ensuring an apples-to-apples comparison with our PyTorch training loop.  The same schedule is applied unmodified to all datasets and architectures, reflecting the common practitioner practice of adopting AutoAugment "as is".

\paragraph{ModelAuditor}
When the auditor recommended a custom augmentation pipeline, we first mapped each suggested operation to its closest standard transform in the torchvision v2 API.  If no equivalent existed (e.g.\ histogram equalization), we wrapped the agent-supplied Python snippet into a torchvision-compatible transform class. The resulting pipeline, automatically specific to both dataset and architecture, replaced AutoAugment in the training dataloader.

\subsection{Code Availability} 
We release the core model auditor package with a unified metrics and shift interface \url{https://github.com/MLO-lab/ModelAuditorCore/} as well as the main agentic command line tool  \url{https://github.com/MLO-lab/ModelAuditor/}.

\subsection{Data Availability} 
The Camelyon17 dataset is available at \url{https://wilds.stanford.edu/datasets/}. HAM10000 can be found here \url{https://dataverse.harvard.edu/dataset.xhtml?persistentId=doi:10.7910/DVN/DBW86T}. CheXpert and ChestXray14 can be found here \url{https://stanfordmlgroup.github.io/competitions/chexpert/} and here \url{https://www.kaggle.com/datasets/nih-chest-xrays/data} respectively.

The SIIM-ISIC model weights can be found here \url{https://zenodo.org/doi/10.5281/zenodo.10049216}. All ISIC images can be accessed under \url{https://challenge.isic-archive.com/data}.

\bmhead{Acknowledgements}
This work was co-funded by the European Union (ERC, TAIPO, 101088594 to F.B.). Views and opinions expressed are those of the authors only and do not necessarily reflect those of the European Union or the ERC. Neither the European Union nor the granting authority can be held responsible for them.

\bibliography{sn-bibliography}

\appendix

\newpage



\begin{appendices}
\section{Extended Results}
\begin{table*}[hb]
\centering
\resizebox{\textwidth}{!}{%
\renewcommand{\arraystretch}{1.2}
\begin{tabular}{llcccccc}
\hline
\multicolumn{2}{c}{} &
\multicolumn{2}{c}{\shortstack{\rule{0pt}{2.5ex}Balanced \\ Accuracy ↑}} &
\multicolumn{2}{c}{\shortstack{Matthews \\ Corr. ↑}} &
\multicolumn{2}{c}{\shortstack{Brier \\ Score ↓}} \\
\textbf{Model} & Variant & ID & OOD & ID & OOD & ID & OOD \\
\hline
\multirow{3}{*}{ResNet50}
& Base           & 98.88 $\pm$ 0.06 & 87.05 $\pm$ 2.91 & \textbf{98.05} $\pm$ 0.50 & 74.57 $\pm$ 5.91 & \textbf{1.91} $\pm$ 0.12 & 20.46 $\pm$ 3.00 \\
& AutoAugment    & 98.53 $\pm$ 0.07 & 91.81 $\pm$ 0.99 & 97.05 $\pm$ 0.15 & 84.40 $\pm$ 1.71 & 2.37 $\pm$ 0.08 & 13.82 $\pm$ 1.82 \\
& ModelAuditor   & \textbf{98.75} $\pm$ 0.06 & \textbf{95.28} $\pm$ 0.94 & 97.50 $\pm$ 0.12 & \textbf{90.61} $\pm$ 1.83 & 1.93 $\pm$ 0.07 & \textbf{7.66} $\pm$ 1.44 \\
\hline
\multirow{3}{*}{ResNet152} 
& Base          & \textbf{98.95} $\pm$ 0.06 & 88.49 $\pm$ 1.87 & \textbf{97.90} $\pm$ 0.11 & 77.60 $\pm$ 3.43 & \textbf{1.75} $\pm$ 0.09 & 19.43 $\pm$ 3.64 \\
& AutoAugment   & 98.79 $\pm$ 0.11 & 93.45 $\pm$ 0.87 & 97.58 $\pm$ 0.22 & 87.31 $\pm$ 1.59 & 1.91 $\pm$ 0.09 & 11.42 $\pm$ 1.59 \\
& ModelAuditor  & 98.85 $\pm$ 0.04 & \textbf{94.88} $\pm$ 0.61 & 97.71 $\pm$ 0.07 & \textbf{89.75} $\pm$ 1.22 &  \textbf{1.75} $\pm$ 0.06 & \textbf{8.05} $\pm$ 1.01 \\
\hline
\multirow{3}{*}{VGG19} 
& Base          & \textbf{98.92} $\pm$ 0.01 & 95.18 $\pm$ 0.48 & \textbf{97.85} $\pm$ 0.02  & 90.71 $\pm$ 1.02 & \textbf{1.92} $\pm$ 0.04 & 8.57 $\pm$ 1.02 \\
& AutoAugment   & 98.74 $\pm$ 0.05 & 96.21 $\pm$ 0.14  & 97.47 $\pm$ 0.10 & 92.51 $\pm$ 0.32 & 2.01 $\pm$ 0.02 & 6.29 $\pm$ 0.21 \\
& ModelAuditor  & 98.67 $\pm$ 0.07 & \textbf{96.25} $\pm$ 0.10 & 97.37 $\pm$ 0.15 & \textbf{92.57} $\pm$ 0.23 & 2.14 $\pm$ 0.01 & \textbf{5.93} $\pm$ 0.13 \\
\hline
\multirow{3}{*}{ViT} 
& Base          & \textbf{98.28} $\pm$ 0.10 & 73.76 $\pm$ 2.20 & \textbf{96.58} $\pm$ 0.19 & 52.12 $\pm$ 3.80 & 2.92 $\pm$ 0.16  & 48.23 $\pm$ 4.58 \\
& AutoAugment   & \textbf{98.28} $\pm$ 0.02 & 86.39 $\pm$ 1.05 & 96.57 $\pm$ 0.05 & 74.52 $\pm$ 1.94 & \textbf{2.59} $\pm$ 0.06 & 21.77 $\pm$ 1.55 \\
& ModelAuditor  & 98.11 $\pm$ 0.07 & \textbf{88.44} $\pm$ 2.14 & 96.24 $\pm$ 0.14 & \textbf{78.07} $\pm$ 3.78 & 2.95 $\pm$ 0.08 & \textbf{17.95} $\pm$ 3.35 \\
\hline
\end{tabular}%
}
\caption{Balanced Accuracy, Matthews Correlation Coefficent and Brier Score for the Wilds Camelyon17 In- and Out-of-distribution dataset for ResNet50, ResNet152, VGG19 and ViT with and without ModelAuditor training improvements. Best values for each architecture highlighted in bold.}
\label{tab:camelyon17}
\end{table*}

\begin{figure}[H]
    \centering
    \includegraphics[width=\linewidth]{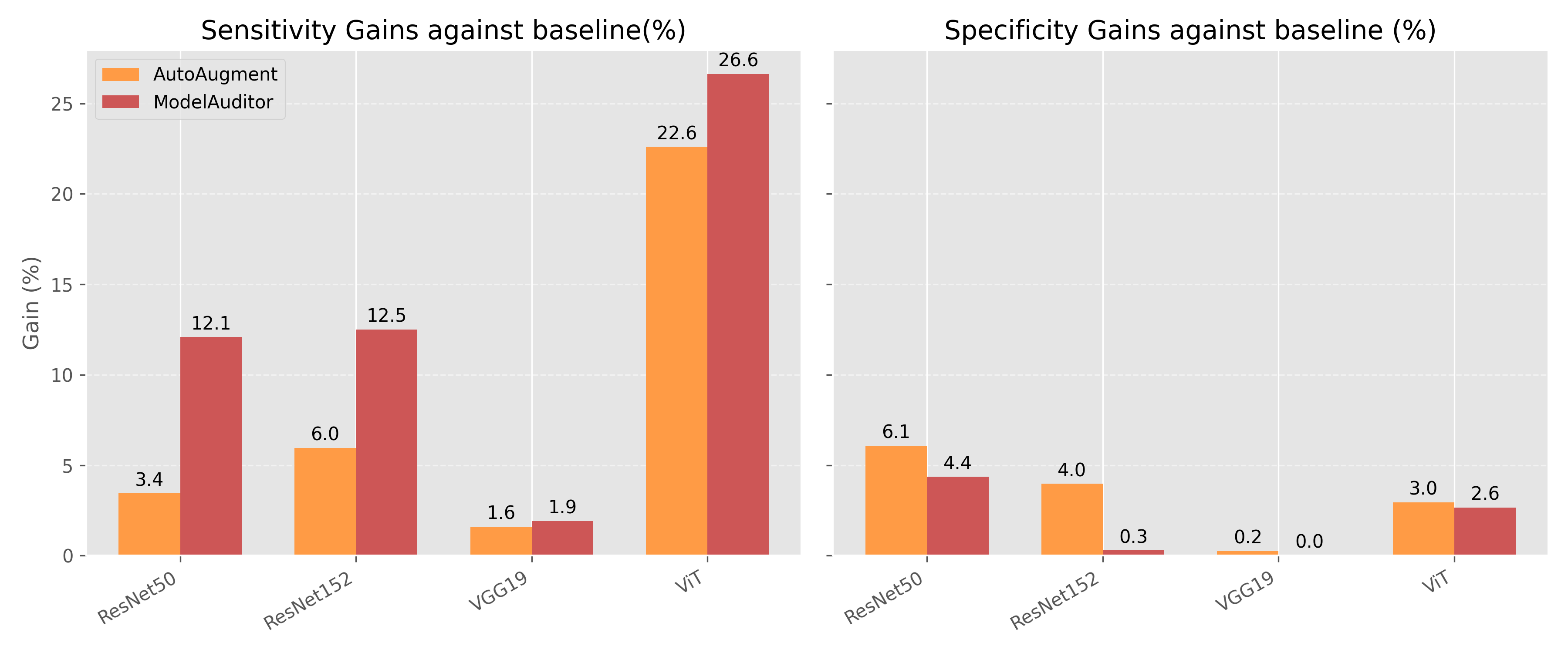}
    \caption{ \textbf{Comparison of Sensitivity and Specificity for ModelAuditor and AutoAugment on Camelyon17:} ModelAuditor consistently outperforms AutoAugment in increasing sensitivity against the base model. A highly sensitive model minimizes false negatives, meaning it detects disease (true positive) more often.}
    \label{fig:comparison}
\end{figure}

\begin{table*}[!ht]
  \centering
  \label{tab:chexpert}
  \begin{tabular}{llcc}
    \toprule
    \textbf{Model} & \textbf{Variant}     & \textbf{AUROC ID ↑} & \textbf{AUROC OOD ↑} \\
    \midrule
    \multirow{3}{*}{ResNet50}   & Base            & 78.62 $\pm$ 0.11 & 74.80 $\pm$ 0.44 \\
                                & AutoAugment     & 80.53 $\pm$ 0.14 & 77.49 $\pm$ 0.09 \\
                                & ModelAuditor    & \textbf{80.85} $\pm$ 0.09 & \textbf{77.68} $\pm$ 0.38 \\
    \midrule
    \multirow{3}{*}{ResNet152}  & Base            & 78.49 $\pm$ 0.00 & 76.22 $\pm$ 0.49 \\
                                & AutoAugment     & \textbf{80.67} $\pm$ 0.09 & 78.84 $\pm$ 0.32 \\
                                & ModelAuditor    & 80.65 $\pm$ 0.13 & \textbf{78.93} $\pm$ 0.10 \\
    \midrule
    \multirow{3}{*}{VGG19}      & Base            & 78.62 $\pm$ 0.03 & 74.05 $\pm$ 0.16 \\
                                & AutoAugment     & 80.60 $\pm$ 0.13 & 75.93 $\pm$ 0.96 \\
                                & ModelAuditor    & \textbf{80.74} $\pm$ 0.16 & \textbf{76.63} $\pm$ 0.18 \\
                                
    \midrule
    \multirow{3}{*}{ViT}        & Base            & 76.70 $\pm$ 0.11 & 71.78 $\pm$ 0.29 \\
                                & AutoAugment     & 79.48 $\pm$ 0.06 & 77.04 $\pm$ 0.09 \\
                                & ModelAuditor    & \textbf{80.31} $\pm$ 0.07 & \textbf{77.26} $\pm$ 0.33 \\
    \bottomrule
  \end{tabular}
  
  \caption{AUROC for In-Distribution (ID) and Out-of-Distribution (OOD) across models and variants for CheXpert and Chest-Xray14. Best values for each architecture highlighted in bold.}
\end{table*}

\begin{table*}[!htb]
  \centering
  \begin{tabular}{llcc}
    \toprule
    \textbf{Model} & \textbf{Variant}     & \textbf{AUROC ↑} & \textbf{Brier Score ↓} \\
    \midrule
    \multirow{3}{*}{ResNet50}   & Base            & 78.73 $\pm$ 2.24 & 47.10 $\pm$ 2.97  \\
                                & AutoAugment     & 74.76 $\pm$ 1.20 & 44.21 $\pm$ 1.80  \\
                                & ModelAuditor    &  \textbf{82.25} $\pm$ 0.26 &  \textbf{36.47} $\pm$ 0.82  \\
    \midrule
    \multirow{3}{*}{ResNet152}  & Base            & 76.06 $\pm$ 2.38 & 51.02 $\pm$ 3.89  \\
                                & AutoAugment     & 77.56 $\pm$ 1.24 &  \textbf{38.91} $\pm$ 1.12  \\
                                & ModelAuditor    &  \textbf{80.99} $\pm$ 1.37 & 38.96 $\pm$ 1.22  \\
    \midrule
    \multirow{3}{*}{VGG19}      & Base            & 78.96 $\pm$ 2.46   & 48.74 $\pm$ 4.26 \\
                                & AutoAugment     & 71.19 $\pm$ 0.88   & 50.58 $\pm$ 2.22 \\
                                & ModelAuditor    &  \textbf{80.28} $\pm$ 0.41   &  \textbf{40.08} $\pm$ 0.76 \\
    \midrule
    \multirow{3}{*}{ViT}        & Base            & 80.08 $\pm$ 0.78  & 37.23 $\pm$ 0.61 \\
                                & AutoAugment     & 68.53 $\pm$ 9.04  & 44.03 $\pm$ 4.03 \\
                                & ModelAuditor    &  \textbf{80.24} $\pm$ 0.83  &  \textbf{36.51} $\pm$ 0.68 \\
    \bottomrule
  \end{tabular}
  \caption{Out-of-distribution (OOD) Performance (AUROC and Brier score) across models and variants for HAM10000. Best values for each architecture highlighted in bold.}
  \label{tab:ham10000}
\end{table*}

\end{appendices}

\newpage
\section{Prompts}
\subsection{Metrics Selection Prompt}
You'll determine appropriate metrics for image classification tasks by analyzing the provided information and following the Metrics Reloaded framework recommendations. You'll extract what's known from the user's input and only ask for missing critical information.

\medskip
\noindent\textbf{Decision Logic from Metrics Reloaded:}

We are only working with image classification tasks and you'll select metrics from these categories:

\medskip
\noindent\textbf{1. Multi-class Counting Metrics:}
\begin{itemize}
    \item When classes are balanced: \texttt{accuracy}
    \item When classes are imbalanced with compensation requested: \texttt{balanced\_accuracy} or \texttt{matthews\_correlation\_coefficient}
    \item When there's unequal severity of class confusions: \texttt{expected\_cost}
\end{itemize}

\medskip
\noindent\textbf{2. Per-class Counting Metrics:}
\begin{itemize}
    \item For binary problems: \texttt{sensitivity} (recall) \texttt{specificity}
    \item When preference for minimizing false positives: \texttt{positive\_predictive\_value} or \texttt{f\_beta\_score} ($\beta < 1$)
    \item When preference for minimizing false negatives: \texttt{negative\_predictive\_value} or \texttt{f\_beta\_score} ($\beta > 1$)
    \item When cost-benefit analysis is needed: \texttt{net\_benefit} or \texttt{adjusted f\_beta\_score}
\end{itemize}

\medskip
\noindent\textbf{3. Multi-threshold Metrics:}
\begin{itemize}
    \item Default recommendation: \texttt{auroc}
    \item When classes are highly imbalanced: \texttt{average\_precision}
\end{itemize}

\medskip
\noindent\textbf{4. Calibration Metrics:}
\begin{itemize}
    \item When calibration assessment is requested: \texttt{expected\_calibration\_error} and \texttt{root\_brier\_score}
    \item For comparison of calibration methods: \texttt{kernel\_calibration\_error}
    \item For overall probabilistic performance: \texttt{negative\_log\_likelihood} or \texttt{brier\_score}
\end{itemize}

\medskip
\hrule
\medskip

\noindent\textbf{You'll look for:}
\begin{itemize}
    \item Binary vs multi-class classification
    \item Class balance vs imbalance
    \item Error preference (false positives vs false negatives)
    \item Need for probability calibration
    \item Decision threshold requirements
\end{itemize}

If any critical information is missing you'll ask targeted questions before providing recommendations. Only end the conversation by outputting the metrics if you know all the important facts!

\medskip
\hrule
\medskip

\noindent\textbf{Output Format}

You'll output a simple list of recommended metric names inside a \texttt{<metric>} tag without hyperparameters or explanations when you are done asking questions:

\medskip
\noindent\textit{Example when asking questions:}

Are the classes in your classification task imbalanced?

\medskip
\noindent\textit{Example when done:}

\begin{verbatim}
<metric>
Accuracy
MatthewsCorrelationCoefficient
Sensitivity
Specificity
AUROC
ExpectedCalibrationError
</metric>
\end{verbatim}

\medskip
\hrule
\medskip

\noindent\textbf{Available Metrics}

Here is a list of all available metrics and how you should call them:

\texttt{Sensitivity}, \texttt{Specificity}, \texttt{PositivePredictiveValue}, \texttt{NegativePredictiveValue}, \texttt{PositiveLikelihoodRatio}, \texttt{DiceSimilarityCoefficient}, \texttt{FBetaScore}, \texttt{NetBenefit}, \texttt{Accuracy}, \texttt{BalancedAccuracy}, \texttt{MatthewsCorrelationCoefficient}, \texttt{WeightedCohensKappa}, \texttt{ExpectedCost}, \texttt{AUROC}, \texttt{AveragePrecision}, \texttt{BrierScore}, \texttt{RootBrierScore}, \texttt{ExpectedCalibrationError}, \texttt{ClassWiseECE}, \texttt{ECEKernelDensity}, \texttt{KernelCalibrationError}, \texttt{NegativeLogLikelihood}

\medskip
\hrule
\medskip

\noindent\textbf{Important!} When dealing with Multi-Label classification (so multiple labels can be true), we can only select from the following special metrics:

\texttt{MultiLabelSubsetAccuracy}, \texttt{MultiLabelHammingLoss}, \texttt{MultiLabelPrecision}, \texttt{MultiLabelRecall}, \texttt{MultiLabelF1Score}, \texttt{MultiLabelJaccardScore}, \texttt{MultiLabelAUROC}

\medskip
\hrule
\medskip

Ask questions step by step without numbering and not all at once. Only output the metric tag list and nothing else when you are done!

\subsection{Shifts Prompt}
You now should also decide which distribution shifts should be used to test the data out of distribution. Do that based on the previous descriptions of the use case.

If any critical information is missing you'll ask targeted questions before providing recommendations. Only end the conversation by outputting the metrics if you know all the important facts!

\medskip
\hrule
\medskip

\noindent\textbf{Output Format}

You'll output a list of recommended distribution shifts inside a \texttt{<shift>} tag. For shifts that support parameters, include the values in Python list format after the shift name.

\medskip
\noindent\textit{Example when asking questions:}

Will you use your model under different lighting settings?

\medskip
\noindent\textit{Example when done:}

\begin{verbatim}
<shift>
GaussianNoise([0, 0.05, 0.1])
BrightnessShift([0.8, 1.2, 1.5])
Rotation([90, 180, 270])
HorizontalFlip
</shift>
\end{verbatim}

\medskip
\hrule
\medskip

\noindent\textbf{Available Shifts and Parameters}

Here is a list of all available shifts and their parameters:

\begin{itemize}
    \item \texttt{GaussianNoise([factor\_values])} - Adds Gaussian noise with the specified factor
    \item \texttt{BrightnessShift([factor\_values])} - Adjusts brightness by the specified factors ($>1$ brightens, $<1$ darkens)
    \item \texttt{Rotation([degrees\_values])} - Rotates the image by specified degrees
    \item \texttt{HorizontalFlip} - Flips the image horizontally (no parameters)
    \item \texttt{VerticalFlip} - Flips the image vertically (no parameters)
    \item \texttt{ContrastShift([factor\_values])} - Adjusts contrast by the specified factors ($>1$ increases, $<1$ decreases)
    \item \texttt{SaturationShift([factor\_values])} - Adjusts saturation by the specified factors ($>1$ increases, $<1$ decreases)
    \item \texttt{HueShift([factor\_values])} - Adjusts hue by the specified factors (range typically 0.0-0.25)
    \item \texttt{GaussianBlur([kernel\_size\_values])} - Applies Gaussian blur with specified kernel sizes (odd numbers)
    \item \texttt{JPEGCompression([quality\_values])} - Applies JPEG compression with specified quality values (lower = more compression)
    \item \texttt{Pixelation([scale\_values])} - Reduces image resolution then upsamples (lower values = more pixelation)
    \item \texttt{PerspectiveTransform([distortion\_values])} - Applies perspective distortion with increasing intensity
    \item \texttt{ZoomIn([scale\_values])} - Zooms in with specified scale factors (lower values = more zoom)
    \item \texttt{ZoomOut([scale\_values])} - Zooms out with specified scale factors (lower values = more zoom out)
    \item \texttt{RandomErasing([area\_values])} - Erases a random rectangle with specified area proportions
    \item \texttt{Grayscale([intensity\_values])} - Converts to grayscale with specified intensity (1.0 = full grayscale)
    \item \texttt{Sharpness([factor\_values])} - Adjusts sharpness by specified factors ($>1$ increases, $<1$ decreases)
    \item \texttt{ColorJitter([magnitude\_values])} - Applies simultaneous brightness, contrast, saturation, and hue shifts
    \item \texttt{ElasticTransform([alpha\_values])} - Applies elastic deformation with specified intensities
    \item \texttt{Solarize([threshold\_values])} - Inverts pixels above threshold (lower values = more solarization)
    \item \texttt{Posterize([bits\_values])} - Reduces color bits per channel (lower values = more posterization)
    \item \texttt{MotionBlur([kernel\_size\_values])} - Applies motion blur with specified kernel sizes (odd numbers)
\end{itemize}

\medskip
\hrule
\medskip

Ask questions step by step without numbering and not all at once. Only output the shift tag list when you are done!

\subsection{Analysis Prompt}
You received the shift results and should now elaborate the most important points to the user. Only include tables or results if you feel that it will significantly increase the understanding of the user.

Give concrete advice on where the model could potentially fail and how to possibly improve the results.

\medskip
\hrule
\medskip

Give concrete recommendations for transformations that should be implemented in training to mitigate these issues.

\medskip
\noindent\textbf{POSSIBLE TRANSFORMATIONS:}

\medskip
\noindent\textbf{Resize \& Crop}
\begin{itemize}
    \item \texttt{v2.Resize(size[, interpolation, max\_size])}
    \item \texttt{v2.ScaleJitter(target\_size[, scale\_range])}
    \item \texttt{v2.RandomResize(min\_size, max\_size)}
    \item \texttt{v2.RandomShortestSize(min\_size[, max\_size])}
    \item \texttt{v2.RandomCrop(size[, padding])}
    \item \texttt{v2.RandomResizedCrop(size[, scale, ratio])}
    \item \texttt{v2.RandomIoUCrop([min\_scale, max\_scale])}
    \item \texttt{v2.CenterCrop(size)}
    \item \texttt{v2.FiveCrop(size)} → 5 crops (4 corners + center)
    \item \texttt{v2.TenCrop(size[, vertical\_flip])} → 5 crops + flipped
\end{itemize}

\medskip
\noindent\textbf{Flip, Pad, Rotate, Transform}
\begin{itemize}
    \item \texttt{v2.RandomHorizontalFlip([p])}
    \item \texttt{v2.RandomVerticalFlip([p])}
    \item \texttt{v2.Pad(padding[, fill, padding\_mode])}
    \item \texttt{v2.RandomZoomOut([fill, side\_range, p])}
    \item \texttt{v2.RandomRotation(degrees)}
    \item \texttt{v2.RandomAffine(degrees[, translate, scale, shear])}
    \item \texttt{v2.RandomPerspective([distortion\_scale, p])}
    \item \texttt{v2.ElasticTransform([alpha, sigma])}
\end{itemize}

\medskip
\noindent\textbf{Color \& Noise}
\begin{itemize}
    \item \texttt{v2.ColorJitter([brightness, contrast, saturation, hue])}
    \item \texttt{v2.RandomChannelPermutation()}
    \item \texttt{v2.RandomPhotometricDistort([brightness, contrast, ...])}
    \item \texttt{v2.Grayscale([num\_output\_channels])}
    \item \texttt{v2.RGB() / v2.RandomGrayscale([p])}
    \item \texttt{v2.GaussianBlur(kernel\_size[, sigma])}
    \item \texttt{v2.GaussianNoise([mean, sigma, clip])}
    \item \texttt{v2.RandomInvert([p])}
    \item \texttt{v2.RandomPosterize(bits[, p])}
    \item \texttt{v2.RandomSolarize(threshold[, p])}
    \item \texttt{v2.RandomAdjustSharpness(sharpness\_factor[, p])}
    \item \texttt{v2.RandomAutocontrast([p])}
    \item \texttt{v2.RandomEqualize([p])}
\end{itemize}

\medskip
\noindent\textbf{Composition}
\begin{itemize}
    \item \texttt{v2.Compose([...])}
    \item \texttt{v2.RandomApply([...], p)}
    \item \texttt{v2.RandomChoice([...])}
    \item \texttt{v2.RandomOrder([...])}
\end{itemize}

\medskip
\noindent\textbf{Miscellaneous}
\begin{itemize}
    \item \texttt{v2.Normalize(mean, std)}
    \item \texttt{v2.RandomErasing([p, scale, ratio, value])}
    \item \texttt{v2.Lambda(lambd)}
    \item \texttt{v2.SanitizeBoundingBoxes([min\_size])}
    \item \texttt{v2.ClampBoundingBoxes()}
    \item \texttt{v2.UniformTemporalSubsample(num\_samples)}
    \item \texttt{v2.JPEG(quality)}
\end{itemize}

Randomly applying transformations instead of always applying them increases results, so use \texttt{RandomApply} often. For example: \texttt{v2.RandomApply([v2.JPEG(quality=80)], 0.3)}

\medskip
\hrule
\medskip

\noindent\textbf{Analysis Structure}

Answer in the following order:
\begin{enumerate}
    \item List and explain failure cases as well as good results of the model. You don't need to list a lot of different numbers, just the ones that are important.
    \item Explain how this could impact real-world deployment.
    \item Give a simple compose list of transformations that could help mitigate this in training.
    \item Any other information you want to convey to the user.
\end{enumerate}

After the initial advice, answer questions in a concise but helpful manner.

\end{document}